\documentclass[11pt,a4paper]{article}
\usepackage[bookmarks=false]{hyperref}
\usepackage[hyperref]{acl2020}
\usepackage{amsmath}
\usepackage{times}
\usepackage{latexsym}
\usepackage{enumitem}
\usepackage{wrapfig}
\usepackage{multirow}
\usepackage{makecell}
\usepackage{dblfloatfix}
\usepackage{booktabs}

\usepackage{url}
\usepackage{enumitem}
\usepackage{graphicx}

\usepackage{microtype}

\aclfinalcopy 
\def\aclpaperid{***} 

\title{Assessing Effectiveness of Using Internal Signals for Check-Worthy Claim Identification in Unlabeled Data for Automated Fact-Checking}

\author{Archita Pathak \\
  University at Buffalo (SUNY) \\
  New York, USA \\
  \texttt{architap@buffalo.edu} \\ \And
  Rohini K. Srihari \\
  University at Buffalo (SUNY) \\
  New York, USA \\
  \texttt{rohini@buffalo.edu} \\}

\date{}

\begin{document}
\maketitle
\begin{abstract}
While recent work on automated fact-checking has focused mainly on verifying and explaining claims, for which the list of claims is readily available, identifying check-worthy claim sentences from a text remains challenging. Current claim identification models rely on manual annotations for each sentence in the text, which is an expensive task and challenging to conduct on a frequent basis across multiple domains. This paper explores methodology to identify check-worthy claim sentences from fake news articles, irrespective of domain, without explicit sentence-level annotations. We leverage two internal supervisory signals - headline and the abstractive summary - to rank the sentences based on semantic similarity. We hypothesize that this ranking directly correlates to the check-worthiness of the sentences. To assess the effectiveness of this hypothesis, we build pipelines that leverage the ranking of sentences based on either the headline or the abstractive summary. The top-ranked sentences are used for the downstream fact-checking tasks of evidence retrieval and the article's veracity prediction by the pipeline. Our findings suggest that the top 3 ranked sentences contain enough information for evidence-based fact-checking of a fake news article. We also show that while the headline has more gisting similarity with how a fact-checking website writes a claim, the summary-based pipeline is the most promising for an end-to-end fact-checking system.
\end{abstract}

\section{Introduction}

With the rise of social media in recent years, it has become possible to disseminate fake news to millions of people easily and quickly. An MIT media lab study \cite{vosoughi2018spread} from two years ago showed that false information goes six times farther and spreads much faster than real information. Additionally, personalization techniques have enabled targetting people with specific types of fake news based on their interests and confirmation biases. In response, there has been an increase in the number of fact-checking organizations that manually identify check-worthy claims and correct them based on evidence \cite{graves2016rise}. However, a study shows that 50\% of the lifetime spread of some very viral fake news happens in the first 10 minutes, which limits the ability of manual fact-checking - a process that takes a day or two, sometimes a week\footnote{https://www.technologyreview.com/2021/05/03/1023908/machine-learning-project-takes-aim-at-disinformation/}. Automating any part of the fact-checking process can help scale up the fact-checking efforts. Additionally, end-to-end automation can also enable human fact-checkers to devote more time to complex cases that require careful human judgment \cite{Konstantinovskiy2021claimdetection}.

End-to-end automated fact-checking systems involve three core objectives - (1) identifying check-worthy claims, (2) verifying claims against authoritative sources, and (3) delivering corrections/ explanations on the claims \cite{graves2018understanding}. The majority of the recent work focuses on verification and explanation objectives for which a list of claims is readily available \cite{thorne-etal-2018-fever,thorne-vlachos-2018-automated,augenstein-etal-2019-multifc,atanasova-etal-2020-generating,kazemi-etal-2021-extractive}. Identifying check-worthy claims, which is a critical first step for fact-checking, remains a challenging task. 

ClaimBuster is the first work to target check-worthiness \cite{hassan2017kdd}. It is trained on transcripts of 30 US presidential elections debates. Each sentence of the transcripts is annotated for three categories - non-factual sentence, unimportant factual sentence, and check-worthy factual sentence. They then build classifiers to classify sentences into these three labels. Another classification-based approach is to predict whether the content of a given statement makes "an assertion about the world that is checkable" \cite{Konstantinovskiy2021claimdetection}. This approach utilizes annotations for sentences extracted from subtitles of UK political shows. The models are trained to classify statements into binary labels - claim or non-claim. Finally, a system called ClaimRank \cite{jaradat-etal-2018-claimrank} aims to prioritize the sentences that fact-checkers should consider first for fact-checking. ClaimRank is trained on pre-existing annotations on political debates from 9 fact-checking organizations. This approach first classifies the statement as check-worthy or not. The statements are then ranked based on the probabilities that the model assigns to a statement to belong to the positive class. 

While these works are fundamental towards approaching the problem of check-worthy claim identification, the focus is only on a single domain (politics). Additionally, the models rely on sentence-level human annotations, which is an expensive task, challenging to conduct regularly for multiple domains, and subject to personal bias. In this work, we try to overcome these limitations by exploring the effectiveness of using internal signals from unlabeled data. We focus on fake news articles and experiment with two types of internal signals for overall supervision - headline and abstractive summary. We make two hypotheses regarding these signals - first, these two elements of an article contain the gist of the content. To support this hypothesis, we evaluate the headline and the abstractive summary against the manually written Snopes claims for news articles. Claims that Snopes write contain the salient factual idea of the source article. Second, sentences that are semantically close to the headline or the summary are check-worthy. To assess this hypothesis, we experiment with end-to-end fact-checking pipelines. The pipelines leverage the top-ranked sentences relevant to the headline/summary for the downstream fact-checking tasks of evidence retrieval and veracity prediction. The dataset used for these experiments contains articles from multiple domains, such as medical, crime, politics, technology. Through comparative experiments, we find that the top-3 ranked sentences contain enough information for evidence-based fact-checking of a fake news article. We also observe that the summary-based pipeline is the most promising for an end-to-end automated fact-checking system \footnote{Code and data will be available on github upon acceptance.}.

\section{Headline and Summary for Check-Worthiness Identification}
\textbf{Hypothesis: }\textit{The headline and the abstractive summary of the content contain the gist of the article.}

\begin{table}
  \caption{ROUGE scores of generated summaries and headline with the manually written Snope's claim (sample size = 1000 news article)}
  \label{tab:rouge}
  \begin{tabular}{lcc}
    \toprule
    Model & ROUGE-1 & ROUGE-L\\
    \midrule
    BART summary & 9.70 & 70.03\\
    PEGASUS summary & 9.77 & 7.32\\
    T5 summary & 17.71 & 13.34\\
    Headline & \textbf{19.02} & \textbf{16.46}\\
  \bottomrule
\end{tabular}
\end{table}

To analyze this hypothesis, we leverage Snopes dataset\footnote{\url{https://www.kaggle.com/liberoliber/onion-notonion-datasets?select=snopes_phase2_clean_2018_7_3.csv}} which contains metadata relevant to news articles on a variety of topics and Snope's verdict on them. Snope is a popular fact-checking website that has been working for debunking a wide variety of online fake news since 1994\footnote{\url{https://www.snopes.com/about/}}. A typical Snope's verdict usually consists of a manually written ``Claim'' which mentions the main gist of a news article, a veracity rating, and the origin of the claim. We sample 1000 articles from Snope's dataset and calculate ROUGE-1 and ROUGE-L scores to measure the similarity of headline and abstractive summary with manually written Snope's claim. These scores are presented in Table-\ref{tab:rouge}. For abstractive summary generation, we compare the results on three state-of-the-art transformer-based models - BART \cite{lewis-etal-2020-bart}, PEGASUS \cite{pmlr-v119-zhang20ae}, and T5 \cite{JMLR:v21:20-074}. Leveraging Huggingface transformer library \cite{wolf-etal-2020-transformers}, we use ``bart-large-cnn'' model and tokenizer for BART implementation; for PEGASUS implementation, we use ``pegasus-reddit-tifu''; and for T5 we have used ``t5-base''. The generated summary consists of 60-180 tokens. 

While abstractive summary is expected to capture the salient ideas of the article's content, the scores in Table-\ref{tab:rouge} show that the headline has the most gisting similarity with the manually written Snope's claim. This could also imply that Snope's claim resembles the article's headline, probably because the readers mostly remember the headline when searching for fact-checks on Snopes. Nevertheless, since the scores indicate that the headline contains some amount of gist of the article, we include it for further downstream experiments. Among various summaries, the T5 summary outperforms others on this metric. Therefore, we select the T5 summary and the headline as the two approaches to identify check-worthy sentences from a news article.

\section{FACSYS: An End-to-end Fact-Checking System}
\textbf{Hypothesis: }\textit{Sentences that are semantically close to the headline or summary are check-worthy and can be used for fact-checking.}

\begin{figure*}[ht!]
  \centering
  \includegraphics[width=\linewidth]{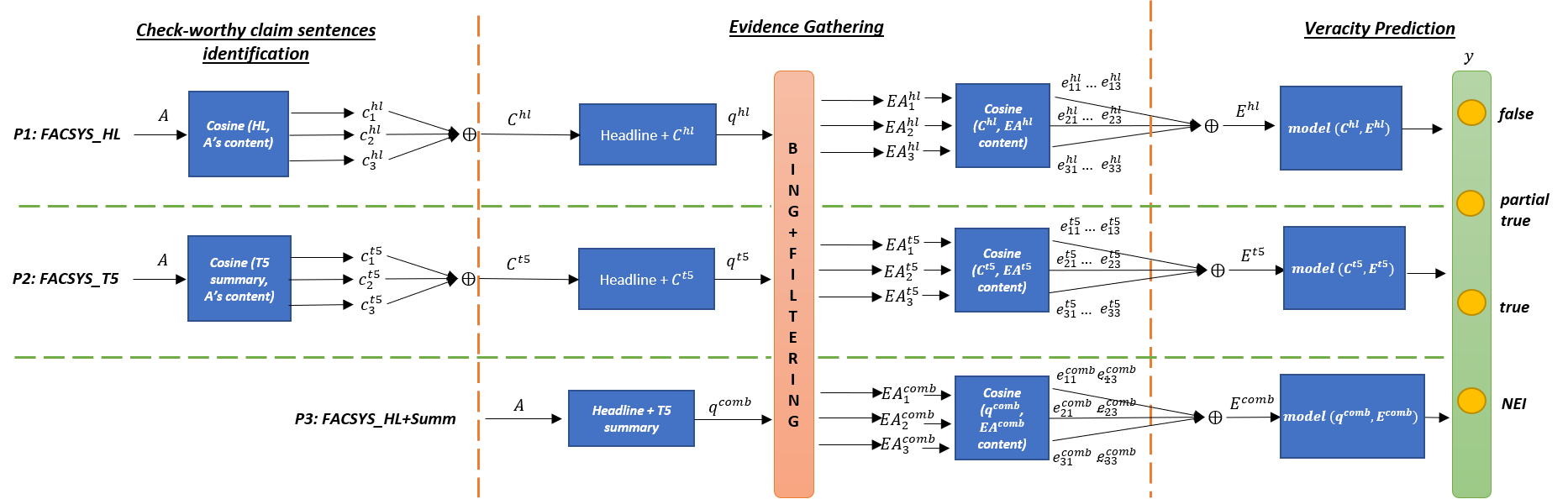}
  \caption{End-to-end fact-checking pipelines used to test the hypothesis: sentences that are semantically close to the headline or summary are check-worthy and can be used for fact-checking an article.}
  \label{fig:pipelines}
\end{figure*}
To test this hypothesis, we build three separate end-to-end fact-checking pipelines - \textbf{P1: FACSYS\_HL}: check-worthy claim sentences are retrieved based on the headline of the article. \textbf{P2: FACSYS\_T5}: check-worthy claim sentences are retrieved based on the T5 summary of the article. \textbf{P3: FACSYS\_HL+Summ}: headline and summary are concatenated and used directly for the downstream fact-checking tasks. The P3 pipeline is created to check the effectiveness of using only the gist of the article for the fact-checking process and comparing it with the other two pipelines. As shown in Figure-\ref{fig:pipelines}, the pipelines consist of the following three stages:

\textbf{1. Check-worthy claim sentences identification:} Given an article $A$ containing set of sentences $s_i$ where $i\in[1, n]$, we first use S-BERT\cite{reimers-2020-multilingual-sentence-bert} to encode $s_i$ for all $i$, and the internal signal headline/summary. We then rank the sentences $s_i$ by calculating the cosine distance between the encoding of each sentence and the internal signal - the lesser the distance, the better the ranking. Through heuristic analyses of the content's quality, we use the top 3 ranked sentences, $\{c_1, c_2, c_3\}$, as check-worthy claim sentences. We observed the relevance of the sentences to the supervisory signals reduces for lower ranked sentences which may introduce noise in the model.

\begin{equation}
\begin{aligned}
    \{(c_1, c_2, c_3\} = Rank(sim_c \{ SBERT(IS), \\
    SBERT(s_i) \}_{i=1}^n)[:3]\}
\end{aligned}
\label{eq:1}
\end{equation}

where $IS \in \{headline, summary\}$ is internal signal. The check-worthy sentences $\{c_1, c_2, c_3\}$ are concatenated as $C$ and supplied as input to the \textit{evidence gathering} stage of the P1 and P2 pipelines.
\begin{equation}
C = c_1 \oplus c_2 \oplus c_3
\end{equation}

For the P3 pipeline, the headline and summary are concatenated and supplied directly to the \textit{evidence gathering} stage.

\textbf{2. Evidence gathering:} This stage retrieves evidence articles $EA$ from the web. It filters them based on the date of publication and the credibility of sources. The query for web search is formulated through simple concatenation:
\begin{equation}
query = headline \oplus C \quad \text{for pipelines P1 and P2,} \quad
\end{equation}
\begin{equation}
query = headline \oplus summary \quad \text{for pipeline P3} \quad
\end{equation}

The final query string is cut short to 40 words due to API limits. We use Bing Search API for firing the query on the web to retrieve the top 35 articles. The filtering of the articles is based on two conditions: (1) the date of publication is within the three months before and after $A$ was published, and (2) the article is from a credible source. For the latter, we use the list of credible sources containing 407 domains rated as ``Least Biased" by Media Bias/Fact Check\footnote{https://mediabiasfactcheck.com/center/}. These sources have been acknowledged by the journalistic community as being factual and well-sourced. We also added a few more left, left-center, right-center, and right news sources to this list based on our own research. For our experiments, we use at most three evidence articles $\{EA_1, EA_2, EA_3\}$, for computational efficiency. Finally, we extract sentences from each evidence article that are most semantically similar to the concatenated check-worthy claim sentences $C$ and rank them using the same process as mentioned in equation \ref{eq:1}. Based on the heuristics used to select top 3 sentences as claim, we also use select top 3 ranked evidence sentences for the next stage.

\textbf{3. Veracity Prediction:} This task is similar to the RTE task on fact-verification of FEVER dataset defined by \cite{thorne-etal-2018-fever}. Given the concatenated check-worthy claim sentences $C$, and three evidence sentences $[e_1, e_2, e_3]$ concatenated as $E$, the main goal is to learn a classifier,
\begin{equation}
g: model(C, E) \rightarrow y
\end{equation}
where $y \in$ \textit{\{ false (0), partial true (1), true (2), NEI (3) \}}

\subsection{Datasets} 
\begin{figure}[h]
  \centering
  \includegraphics[width=\linewidth]{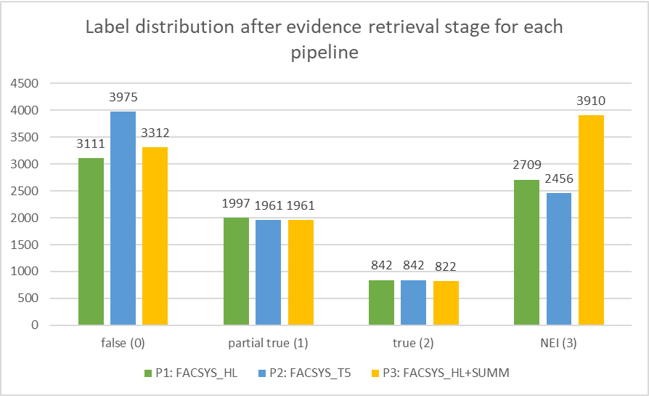}
  \caption{Label distribution of each of the three pipelines for the veracity prediction stage.}
  \label{fig:labels}
\end{figure}

We combine articles from two datasets for fact-checking experiments: (1) Snopes dataset containing articles on various topics published from 1996-2018, and (2) DNF-300 dataset \cite{architap2020} containing 300 articles from the 2016 US presidential election. Both datasets contain news articles along with fact-checking ratings. For the Snopes dataset, we kept only those articles that are labeled as `true', `mostly true', `mixture', `mostly false', `false'. We have grouped `mostly true', `mixture', `mostly false' articles under the single label `partial true' for our experiments. In the DNF-300 dataset, we removed the articles labeled as ``Opinion''.
Additionally, articles for which no evidence article was found after \textit{evidence gathering} stage were labeled as NEI.  Figure-2 shows the label distribution for articles after the \textit{evidence gathering} stage for each of the three pipelines. Most articles for which no evidence was found (labeled as NEI) are for pipeline P3, showing that the quality of query that contains keywords from the abstractive summary is not good. Having keywords from check-worthy claim sentences helps gather good evidence articles as indicated by the NEI label of pipelines P1 and P2. 

\begin{table*}[h]
\centering
\caption{Evaluation of fact-checking pipelines for veracity prediction. Summary based pipeline (pipeline 2) outperforms all other pipelines using a simple BERT\_Concat model.}
\label{tab:results}
\begin{tabular}{llll}
\toprule
\textbf{Pipelines}                   & \textbf{Models}           & \textbf{F1}    & \textbf{LA}    \\
\midrule
Baseline   (Content-Classification)  & BERT                      & 43.57          & 47.74          \\
\midrule
\multirow{3}{*}{P1:  FACSYS\_HL}      & BERT\_Concat               & 64.86          & 68.3           \\
                                     & KGAT (CorefBERT\textsubscript{BASE})     & 60.04          & 63.39          \\
                                     & KGAT (CorefRoBERTa\textsubscript{LARGE}) & 64.08          & 66.55          \\
\midrule
\multirow{3}{*}{P2: FACSYS\_T5}       & BERT\_Concat               & \textbf{67.31} & \textbf{70.34} \\
                                     & KGAT (CorefBERT\textsubscript{BASE})     & 61.65          & 65.43          \\
                                     & KGAT (CorefRoBERTa\textsubscript{LARGE}) & 62.24          & 67.39          \\
\midrule
\multirow{3}{*}{P3:  FACSYS\_HL+SUMM} & BERT\_Concat               & 51.36          & 54.06          \\
                                     & KGAT (CorefBERT\textsubscript{BASE})     & 44.95          & 50.81          \\
                                     & KGAT (CorefRoBERTa\textsubscript{LARGE}) & 54.08          & 54.44         \\
\bottomrule
\end{tabular}
\end{table*}

\subsection{Models}
For the final veracity prediction stage of the pipeline, we use a simple baseline model as follows: 
\begin{equation}
f: MLP(BERT(n)) \rightarrow y
\end{equation}
where n = 500 words of the content of the article. We use BERT \cite{devlin-etal-2019-bert} to encode the words, and the [CLS] token is supplied to the multi-layer perceptron layer for 4-way softmax classification. Note that this classification is only based on the content of the article, irrespective of the evidence. For evidence-based classification, we use the following models:

\textbf{1. BERT\_Concat \cite{zhou-etal-2019-gear}:}  for each pair $(EA,C)$, this approach concatenates evidence sentences $E$ with $C$. BERT is used to perform the natural language inference (NLI) prediction for the final veracity label. Hence, as per BERT's requirement, a separator is used between $E$ and $C$ for the NLI task.

\textbf{2. KGAT (CorefBERT\textsubscript{BASE}) \cite{ye-etal-2020-coreferential}:} KGAT \cite{liu-etal-2020-fine} conducts a fine-grained graph attention network with kernels, developed for FEVER dataset. While the original KGAT model uses BERT, we use this model with CorefBERT\textsubscript{BASE} in our experiments. CorefBERT \cite{ye-etal-2020-coreferential} is a novel language representation model that can capture the coreferential relations in context. The KGAT with CorefBERT\textsubscript{BASE} has shown a significant improvement over KGAT with BERT\textsubscript{BASE} on the FEVER dataset. 

\textbf{3. KGAT (CorefRoBERTa\textsubscript{LARGE}) \cite{ye-etal-2020-coreferential}:} The current state-of-the-art model on the FEVER dataset is KGAT with CorefRoBERTa\textsubscript{LARGE}. CorefBERT, which incorporates coreference information in distant-supervised pre-training, contributes to verifying if the claim and evidence discuss the same mentions, such as a person or an object.

We use the same configurations as specified in the previous papers for these models. Train, validation, and test sets are created by taking 80:10:10 splits. All models are trained for three epochs.

\subsection{Evaluation Metrics}
The official evaluation metrics\footnote{https://github.com/sheffieldnlp/fever-scorer} for the FEVER RTE task include Label Accuracy (LA) and FEVER score. In our work, since we do not have ground-truth evidence (whether a sentence is an evidence or not), we discard the FEVER score metric and only use classification metric F1 along with LA.

\subsection{Results and Discussion}
As shown in Table-\ref{tab:results}, incorporating evidence improves the classification results by over 20\% as compared to the baseline content-based classification. This suggests that converting the entire article into an embedding might be lossy. Focusing only on the check-worthy claim sentences and relevant evidence helps build embedding well-suited to learn the article's main idea for the classification task. Additionally, results on pipeline P3 show that the evidence articles are not at par as compared to pipelines P1 and P2. The reason behind this could be because P3 uses the generated summary in the query string. The generated summary may not contain the same entities as mentioned in the original text. Hence, leading to substandard quality of web articles in the \textit{evidence gathering} stage. 

Further, since the headline of the article is shown to have the best ROUGE scores with the manually written claims by Snopes (Table-\ref{tab:rouge}), it is expected that the check-worthy sentences identified using the headline will have better performance in the \textit{evidence gathering} and the \textit{veracity prediction} stages. However, the P2 pipeline based on the summary outperforms all other pipelines in predicting veracity labels. This indicates that the T5-based abstractive method generates a high-quality summary that helps in determining better ranking of check-worthy claim sentences from the content of the article. While our evaluation shows that this ranking has potential in automated fact-checking of article without any sentence-level annotations, it can also assist human fact-checkers in identifying important sentences from the article for their manual evaluation. In terms of models, a simple BERT\_Concat model obtains better results than the heavy, graph attention-based models, achieving $\approx$23\% improvement on the baseline F1 score and over 30\% improvement on the baseline LA score.

\section{Conclusion}
We explore identification of check-worthy claim sentences from a news article without any sentence-level annotation. We show experiments leveraging two internal signals - headline and abstractive summary of the article. We test two hypotheses - (1) headline/abstractive summary contains the gist of the article, and (2) sentences of the content semantically relevant to the headline/summary are check-worthy claim sentences. We build fact-checking pipelines for this purpose and show that the check-worthy claim sentences identified based on the summary of the article are adequate for downstream tasks of evidence gathering and veracity prediction of the article. Our experiments use articles ranging on a variety of topics and associated with four veracity labels. For future work, we aim to use the abstractive-summary-based methodology for fact-checking of other types of textual data - online discourse, debate transcripts, etc. We believe that leveraging topic detection along with the summary-based check-worthiness identification can help overcome the issues and biases introduced due to sentence-level manual annotations.

\bibliography{acl2020}
\bibliographystyle{acl_natbib}

\end{document}